\documentclass[sigconf]{acmart}
\usepackage{enumitem}
\usepackage{xspace}
\usepackage{booktabs}
\usepackage{graphicx}

\usepackage{xcolor}

\definecolor{WEIRDBLUE}{RGB}{0,114,178}

\AtBeginDocument{%
  }

\setcopyright{acmlicensed}
\copyrightyear{2018}
\acmYear{2018}
\acmDOI{XXXXXXX.XXXXXXX}
\acmConference[Conference acronym 'XX]{Make sure to enter the correct
  conference title from your rights confirmation email}{June 03--05,
  2018}{Woodstock, NY}
\acmISBN{978-1-4503-XXXX-X/2018/06}

\begin{document}

\title{DesignAgent3D: Interactive 3D Scene Editing via Designer-like Multimodal Reasoning}

\author{Xiujin Liu}
\authornote{Both authors contributed equally to this research.}
\email{jeanliu@umich.edu}
\affiliation{%
  \institution{University of Michigan, Ann Arbor}
  \city{Ann Arbor}
  \state{Michigan}
  \country{USA}
}

\author{Tianyu Yang}
\authornotemark[1]
\email{tyang4@nd.edu}
\affiliation{%
  \institution{University of Notre Dame}
  \city{South Bend}
  \state{Indiana}
  \country{USA}
}

\author{Yilun Zhao}
\email{tyang4@nd.edu}
\affiliation{%
  \institution{Yale University}
  \city{New Haven}
  \state{Connecticut}
  \country{USA}
}

\author{Xiangliang Zhang}
\authornote{Corresponding author.}
\email{xzhang22@nd.edu}
\affiliation{%
  \institution{University of Notre Dame}
  \city{South Bend}
  \state{Indiana}
  \country{USA}
}

\renewcommand{\shortauthors}{Trovato et al.}

\begin{abstract}
Text-guided 3D scene editing provides an intuitive interface for modifying reconstructed environments, but remains difficult because natural language design requests are often semantically underspecified and must be grounded in cluttered 3D scenes. Existing methods typically formulate the task as one-shot conditional generation from a single prompt,  
failing to resolve ambiguous user intents or achieve precise spatial grounding. Consequently, they suffer from severe object localization drift, tracking failure under occlusions, and the notorious multi-view ``sticker effect.'' 
To overcome these limitations, we present DesignAgent3D, an interactive multimodal agentic framework that reformulates 3D scene editing as a designer-like \textbf{Plan--Perceive--Act} paradigm. 
The agent first \textit{plans} by interacting with the user to clarify underspecified design goals, then \textit{perceives} by grounding the intended edit to specific objects or regions in the 3D scene, and finally \textit{acts} by applying controlled visual modifications while preserving scene consistency. The edits are further integrated into the underlying 3D representation, supporting persistent and multi-view consistent novel-view rendering.
Extensive experiments across both NeRF and 3D Gaussian Splatting backbones demonstrate that DesignAgent3D significantly outperforms state-of-the-art baselines, delivering superior semantic intent alignment, impeccable spatial localization accuracy, and high-fidelity multi-view consistency. 
\end{abstract}



\received{20 February 2007}
\received[revised]{12 March 2009}
\received[accepted]{5 June 2009}

\maketitle

\begin{figure}[t]
    \centering
    \includegraphics[width=\columnwidth]{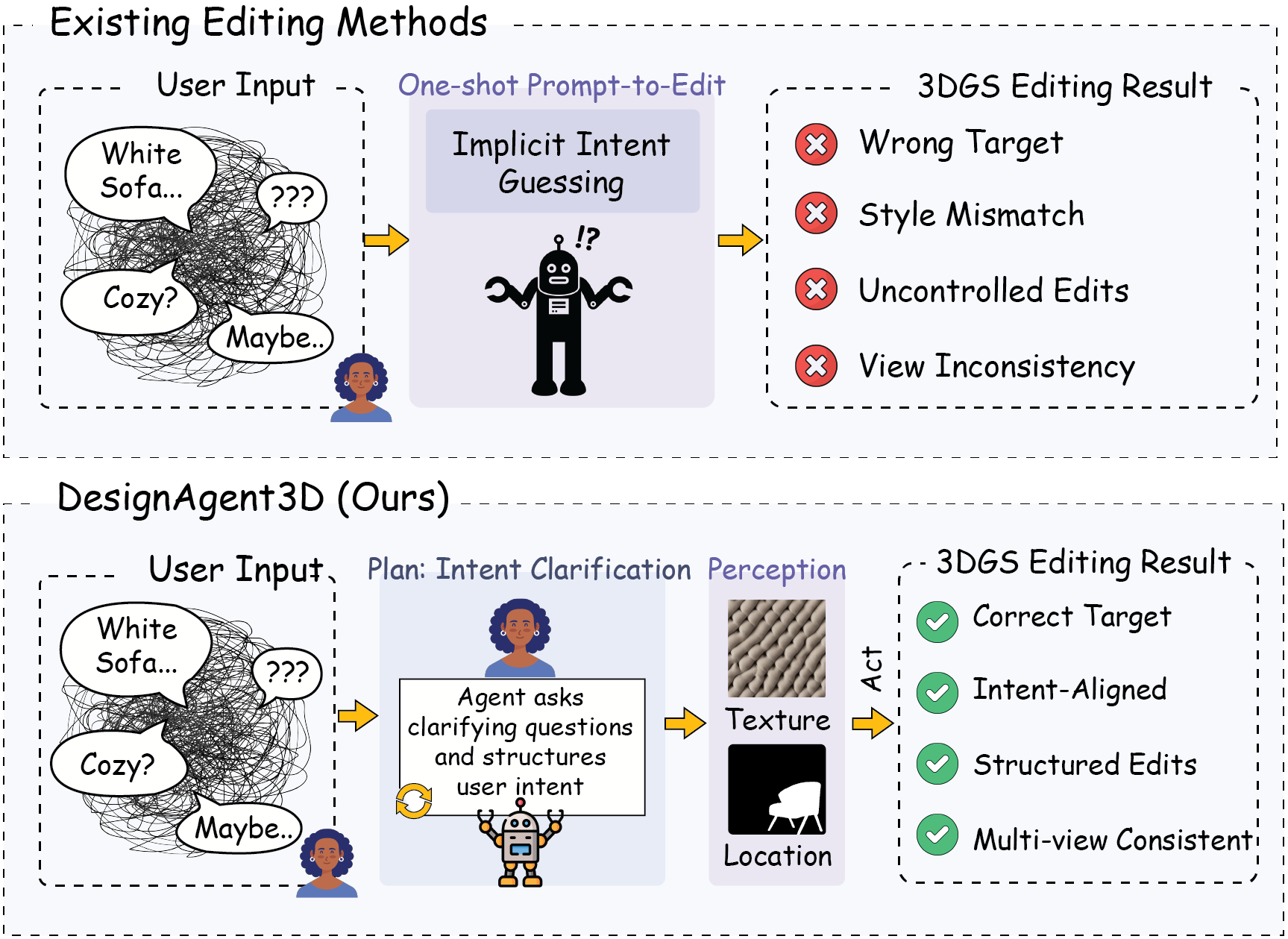} 
    \caption{\textbf{Conceptual comparison between traditional one-shot 3D editing  methods and DesignAgent3D.} Traditional approaches rely on direct and one-shot prompt-to-scene optimization, frequently suffering from object localization drift or background corruption due to semantic underspecification. In contrast, DesignAgent3D adopts a designer-like \textbf{Plan--Perceive--Act} paradigm that clarifies ambiguous intents via human-in-the-loop collaboration and leverages geometric spatial reasoning, ensuring multi-view consistent and precisely intent-aligned 3D editing results.}
    \label{fig:compare}
\end{figure}

\section{Introduction}

With the recent proliferation of Large Language Models (LLMs) and Multimodal Foundation Models, enabling agents to understand and execute natural language instructions in 3D environments has emerged as a frontier in artificial intelligence~\cite{haque2023instruct, chen2024gaussianeditor}. Concurrently, neural rendering advancements like Neural Radiance Fields (NeRF)~\cite{mildenhall2021nerf} and 3D Gaussian Splatting (3DGS)~\cite{kerbl20233d} have enabled high-fidelity 3D scene reconstruction. The convergence of these fields gives rise to the task of \textit{interactive 3D scene design}: empowering an intelligent system to accurately interpret a user's natural language request, ground it to a specific 3D physical entity, and perform localized multi-view consistent modifications (e.g., altering material, texture, or style). This task serves as a crucial testbed for language-driven spatial intelligence, with profound applications in interior design, interactive simulation, and virtual staging.

\begin{figure*}[t]
    \centering
    \includegraphics[width=\textwidth]{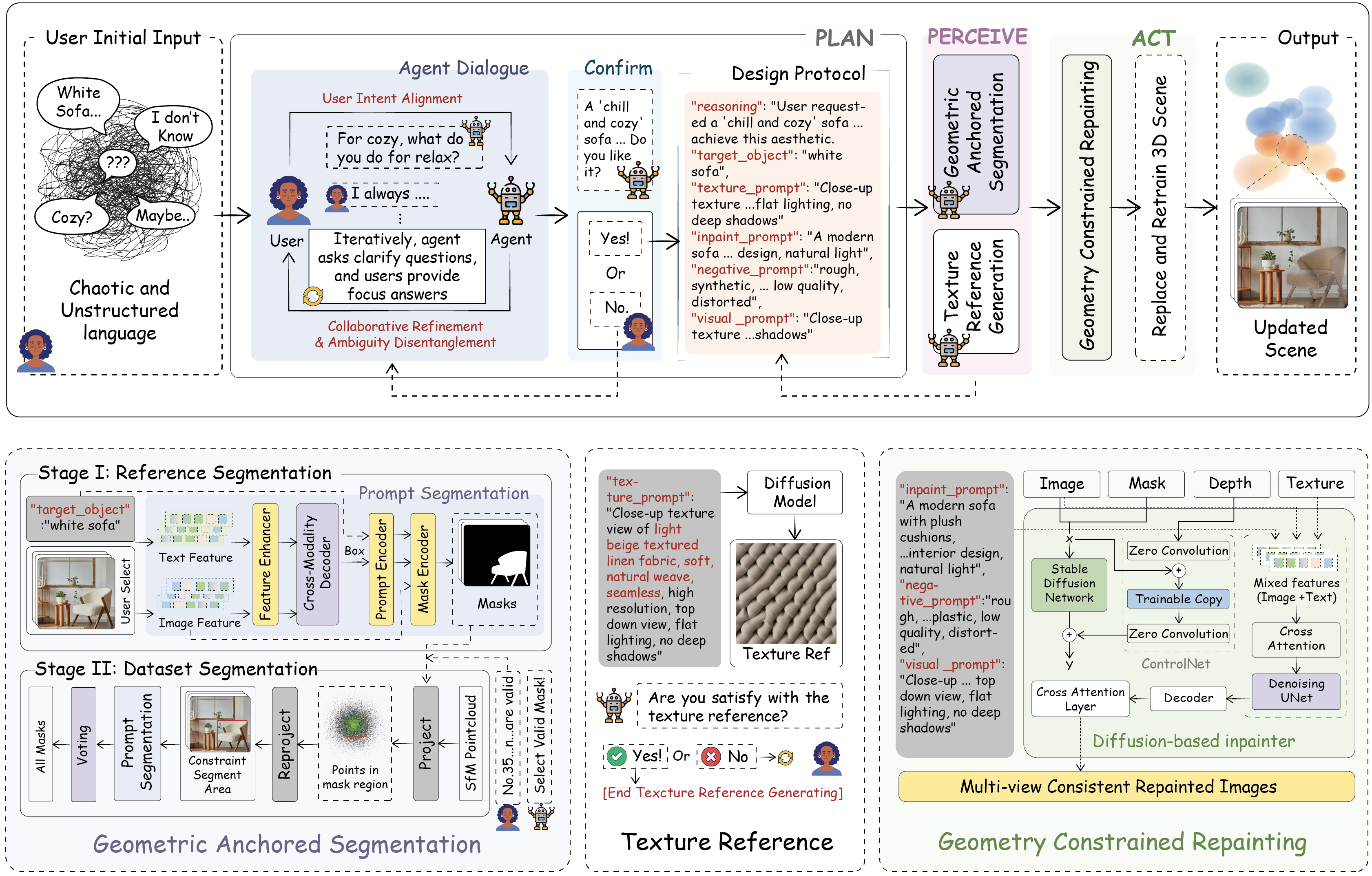} 
    \caption{\textbf{Overview of the DesignAgent3D Pipeline.} The top row illustrates our agentic \textbf{Plan--Perceive--Act} workflow, where the agent clarifies the user request, generates a structured design protocol, grounds the target object and texture, and updates the 3D scene through geometry-constrained repainting. The bottom row details the three core modules: \textbf{Geometric Anchored Segmentation} for view-consistent object masks, \textbf{Texture Reference Generation} for a user-verified material prior, and \textbf{Geometry-Constrained Repainting} for multi-view consistent edited images.}
    \label{fig:Designer_pipeline}
\end{figure*}

The fundamental challenge of this task lies in the inherent nature of human communication: natural language design requests are often abstract, subjective, and semantically underspecified. For instance, when viewing a reconstructed room, a user 
might say, "make this sofa feel warmer," or "change the table to a lighter oak wood."  
Although such instructions are easily interpreted by humans through contextual reasoning and shared visual experience, they are difficult for current AI systems because the intended edits are not explicitly defined in terms of geometry, materials, lighting, or spatial constraints.
To execute these commands reliably, a system must go beyond mapping words to visual changes: it must infer the user’s design intent, identify the referenced objects in a complex 3D layout, understand how material and appearance changes should be applied consistently across surfaces, and preserve the underlying geometry and scene coherence.

Despite recent progress in text-guided 3D editing~\cite{vachha2024instruct, lee2025editsplat}, existing methods largely formulate the task as \emph{one-shot conditional generation}, directly optimizing a 3D representation from a single text instruction. This formulation collapses the full design workflow into one opaque step, forcing the system to infer user intent, locate the target region, and execute the visual modification \emph{simultaneously}. As a result, it suffers from three key limitations. \emph{First, it lacks explicit planning for ambiguous design intent}. For example, an instruction such as ``make the dining area more modern'' does not specify whether the edit should affect the chairs, table, lighting, wall color, or their combination; a one-shot method must nevertheless commit to one interpretation without clarification. \emph{Second, it lacks robust perception for spatial reference grounding.} In a cluttered scene with multiple similar objects, a request such as ``edit the chair next to the cabinet'' requires identifying the exact 3D object instance before editing, yet one-shot methods often rely on weak semantic or view-dependent cues. Without reliable 3D-aware text grounding, localized editing often degenerates into inconsistent multi-view textures, exhibiting the notorious "sticker effect"~\cite{chen2024gaussianeditor}.   These limitations make one-shot generation fundamentally mismatched with real-world interior design, where a  professional human designer never jumps directly to execution from a vague prompt.

To address this gap, we propose an agentic framework for interactive 3D scene editing inspired by the collaborative workflow of human designers: \textbf{Plan -- Perceive -- Act}. Our system first \textbf{\emph{plans}} by engaging with the user in multi-turn clarification to resolve and structure ambiguous user intents, then \textbf{\emph{perceives}} the 3D scene to ground the intended edit to specific objects or regions, and finally \textbf{\emph{acts}} by executing precise visual modifications under strict geometric constraints. 

Following this paradigm, we introduce \textbf{DesignAgent3D}, a multimodal agentic framework that operationalizes interactive 3D scene design through structured human-in-the-loop reasoning. \textbf{DesignAgent3D} systematically decomposes the workflow into three tightly coupled stages: \emph{planning, perception} and \emph{action}.

In the \textbf{Planning} stage, a cognitive dialogic module interacts with the user to resolve semantic ambiguities. It translates vague natural language intents into a structured, machine-executable \textit{Design Protocol}, which explicitly formalizes target object identities, desired visual attributes, style profiles, and negative constraints. This transforms implicit human intent into an explicit, verifiable representation before any rendering pipeline is invoked.

In the \textbf{Perception} stage,  DesignAgent3D grounds this structured linguistic protocol into the physical 3D scene. We propose a \emph{Geometric Anchored Segmentation} module that fuses open-vocabulary semantic segmentation with sparse Structure-from-Motion (SfM) geometry. By propagating sparse spatial priors and incorporating user-validated reference masks, this module successfully solves the multi-view reference resolution problem, distinguishing the target object from visually similar distractors even under severe occlusions. Concurrently, a \emph{Texture Reference Generation} module establishes a canonical material anchor for visual verification.

In the \textbf{Action} stage, the framework performs geometry-constrained visual synthesis. Leveraging the resolved multi-view masks and canonical texture references under depth constraints, DesignAgent3D repaints the target object with cross-view consistency. A replace-and-retrain strategy seamlessly back-propagates these visual edits into the underlying 3D representation (e.g., NeRF \cite{mildenhall2021nerf} or 3DGS \cite{kerbl20233d}).

To demonstrate the effectiveness of \textbf{DesignAgent3D}, we evaluate it on three widely adopted benchmarks including \textit{Nerfstudio} ~\cite{mildenhall2021nerf}, \textit{Mip-Nerf 360}~\cite{barron2022mip}, and \textit{Replica}~\cite{straub2019replica}. Comprehensive quantitative and qualitative comparisons demonstrate that our framework consistently outperforms state-of-the-art one-shot baselines, achieving an order-of-magnitude improvement in text alignment. Notably, DesignAgent3D successfully eliminates the chaotic global distortions and multi-view blurring artifacts typical of prior methods, ensuring precise localized modifications and exceptional background geometric fidelity.


Our contributions are summarized as follows:
\begin{itemize}[leftmargin=*]
\item We reformulate interactive 3D scene editing as a structured \textbf{Plan--Perceive--Act} process, addressing the mismatch between one-shot editing and real-world interior design workflows.
\item We introduce \textbf{DesignAgent3D}, a multimodal agentic framework that clarifies ambiguous user intent, grounds edits in 3D scenes, and executes geometry-preserving modifications.
\item We validate \textbf{DesignAgent3D} through quantitative and qualitative evaluations across both NeRF \cite{mildenhall2021nerf}  and 3DGS \cite{kerbl20233d} backbones, showing that it significantly reduces object localization drift and mask fragmentation while achieving superior multi-view consistency over previous state-of-the-art baselines. 
\end{itemize}

\section{Related Work}

\subsection{3D Scene Editing}
Recent advances in neural rendering, especially by NeRF~\cite{mildenhall2021nerf} and 3DGS~\cite{kerbl20233d}, have fundamentally redefined the boundaries of high-quality 3D scene representation and 3D scene editing. Early learning-based 3D editing works, such as EditNeRF~\cite{liu2021editing}, primarily operates within constrained, pre-trained latent spaces to manipulate global shape and appearance codes, which heavily restricted their generalization to open-world, complex visual environments. To break these topological boundaries, text-driven 3D editing has emerged as a dominant paradigm by leveraging the rich generative priors of large 2D diffusion models. Frameworks like Instruct-NeRF2NeRF~\cite{haque2023instruct} introduces the iterative dataset update pipeline to alternate between 2D image modification and global 3D optimization. Following the arrival of explicit and differentiable 3DGS representations, recent extensions such as Instruct-GS2GS~\cite{vachha2024instruct} and GaussianEditor~\cite{chen2024gaussianeditor} have significantly accelerated this process, enabling localized, high-fidelity semantic editing through Gaussian semantic tracing. More recently, Editsplat~\cite{lee2025editsplat} proposes Multi-view Fusion Guidance (MFG) to incorporate 3D geometric constraints directly into the diffusion process.  PhysGaussian~\cite{xie2024physgaussian} further extends 3DGS editing with physically grounded dynamic simulation. 

However, despite their impressive generative capabilities, these editing pipelines inherently suffer from severe text ambiguity and a lack of precise object-level control. Because they rely on a one-shot textual prompt to optimize the entire scene or a roughly estimated segmentation, they often struggle to break away from their source topology, leading to distorted geometry or severe texture bleeding in multi-object environments. Furthermore, they provide no inherent mechanism to resolve ambiguous or multi-layered user instructions (e.g., "modify the chair" in a cluttered room with multiple distinct chairs). In contrast, our method formulates 3D scene editing as a \emph{Plan -- Perceive -- Act} process. By explicitly decoupling intention clarification from the low-level 3D optimization loop, our framework accurately resolves verbal and spatial ambiguities before any execution occurs, enabling unprecedented fine-grained object control in dense, complex scenes.
\vspace{-0.3cm}
\subsection{Agent-based 3D Generation}

With the rapid evolution of LLMs and VLMs, agentic workflows have been increasingly integrated into 3D reasoning, planning, and spatial synthesis tasks. A prominent branch focuses on procedural 3D generation. For instance, 3D-GPT~\cite{sun20253d} introduces a multi-agent framework that leverages LLMs to dissect text instructions and extract procedural parameters, enabling intuitive, interactive, and automated 3D scene generation within software like Blender, while LandCraft~\cite{liu2026landcraft} proposes an AI-assisted authoring tool that utilizes a coarse-to-fine framework combining LLMs and procedural generation to transform textual descriptions into high-quality, production-ready 3D landscapes with controllable spatial and geographic features. SceneCraft~\cite{yang2024scenecraft} leverages agentic planning to formulate structured scene graphs and optimize layout-aware room scales through multi-view diffusion models. In the field of embodied AI and spatial interaction, frameworks like LEO~\cite{huang2023embodied} demonstrates profound competencies in multi-modal 3D reasoning, continuous navigation, and fine-grained robotic manipulation. Moving towards collaborative systems, contemporary works such as Idea23D~\cite{chen2025idea23d} and DAAO~\cite{su2026difficulty} explore interactive multi-agent cooperation to incrementally refine complex geometry. 
Nevertheless, these methods mainly target scene generation, planning, or high-level control, rather than fine-grained editing of captured 3D scenes. Our work instead focuses on interactive object-level editing that tightly couples agentic reasoning with geometry-aware execution.

\subsection{Multi-view Consistency in 3D Editing}

Maintaining strict multi-view consistency while modifying a 3D scene remains a fundamental challenge, as independent 2D diffusion processes naturally introduce stochastic viewpoint discrepancies, causing structural artifacts, floater accumulations, or severe blurring during 3D reconstruction. To enforce cross-view alignment, earlier NeRF-based editing methods, ViCA-NeRF~\cite{dong2023vica}, introduces text-driven 3D scene editing method that enforces multi-view consistency by combining depth-derived geometric regularization and latent-code alignment from 2D diffusion models.
SPIn-NeRF~\cite{mirzaei2023spin} proposes a novel 3D scene inpainting method that achieves view-consistent object removal in NeRFs by distilling 2D image inpainters into 3D space using a fast-segmented mask, while also introducing a real-world benchmark dataset for evaluation. Within the 3DGS domain, recent literature has heavily focused on designing geometry- or attention-aware constraints to eliminate cross-view discrepancies.GaussCtrl~\cite{wu2024gaussctrl} utilizes a depth-conditioned ControlNet to guide multi-view 2D updates and proposes an attention-based latent code alignment module to unify appearance across perspectives. VcEdit~\cite{wang2024view} ensures multi-view consistency and prevents mode collapse in 3D Gaussian Splatting (3DGS) editing by integrating cross-attention and editing consistency modules directly into the diffusion-based image modification process.  InterGSEdit~\cite{wen2025intergsedit} introduces a CLIP-based semantic consistency selection strategy along with an attention fusion network to establish a 3D geometry-consistent attention prior from user-selected key views.

Although these methods successfully suppress high-frequency visual artifacts during optimization, they mainly enforce consistency during optimization, but do not explicitly resolve ambiguity in target object grounding. If the target object is improperly grounded or tracked across diverse perspectives, even the most stringent mathematical attention constraints will fail. Our framework addresses both issues by combining interactive intent clarification (i.e., \emph{planning}) with geometric anchored object localization for view-consistent editing (i.e., \emph{perception}).


\section{Methodology}

This section first formalizes the interactive object-level 3D editing problem and then details how the proposed \emph{Plan -- Perceive -- Act} pipeline converts ambiguous natural-language intent into geometrically grounded, multi-view consistent scene updates.

\subsection{Problem Definition: Interactive Object-level 3D Scene Editing}
Given a reconstructed 3D scene representation $\mathcal{S}$ (e.g., a 3D Gaussian Splatting scene \cite{kerbl20233d} or a Neural Radiance Field \cite{mildenhall2021nerf}) optimized from a multi-view image dataset $\mathcal{I} = \{I_i\}_{i=1}^N$ with corresponding camera poses, our task   is to perform interactive object-level 3D scene editing according to a user's natural language design request. 
Unlike global scene stylization, object-level editing seeks to modify the appearance, texture, or material of a specific physical target instance $\mathcal{O} \subset \mathcal{S}$ according to the user request, while strictly preserving the background geometry, lighting, and spatial coherence of the remaining scene $\mathcal{S} \setminus \mathcal{O}$. This objective becomes particularly complex in cluttered indoor environments, where intricate spatial arrangements, multi-layer occlusions, and the frequent recurrence of identical object categories (e.g., multiple identical chairs or lamps) fundamentally confound both semantic understanding and geometric localization. To successfully achieve such precise instance-level modifications within complex, cluttered indoor environments, we must address two fundamental requirements.

The first requirement is to resolve the user's editing intent and identify the exact target object. Before modifying an object-level 3D scene, the system must determine both \textit{what} change the user intends and \textit{which} object the change should be applied to. This is nontrivial because natural language design requests are often underspecified, e.g., "make this corner look more modern", and indoor scenes frequently contain multiple visually similar instances, such as repeated chairs, lamps, or cabinets. A one-shot interpretation of such instructions may therefore select the wrong object or impose an unintended style. To avoid premature and arbitrary decisions, the agent must engage in multi-turn interaction with the user, clarify ambiguous preferences and spatial references, determine the target object $\mathcal{O}$, and convert the user's high-level request into a structured editing plan.

The second requirement is \textit{3D-aware target grounding}: after the target object $\mathcal{O}$ is determined, the system must accurately localize and track it across multiple views. This is challenging because object-level editing cannot be performed reliably by independently modifying 2D images. View-wise edits may produce inconsistent textures, inaccurate boundaries, or the ``sticker effect,'' where the edited appearance appears pasted onto images rather than embedded in the 3D scene. The difficulty is further amplified in cluttered indoor environments, where the target object may be partially occluded by furniture, separated into disconnected visible regions, or absent from certain viewpoints. Therefore, the edit must be grounded in the underlying 3D geometry or scene primitives rather than only in 2D image space. By associating the target object $\mathcal{O}$ with its corresponding 3D elements, the system can derive view-specific masks $\mathcal{M}=\{M_i\}_{i=1}^{N}$ that accurately indicate the visible regions of $\mathcal{O}$ in each image. This prevents the edit from leaking into the background $\mathcal{S}\setminus\mathcal{O}$ and enables the edited views $\mathcal{I}'=\{I'_i\}_{i=1}^{N}$ to remain spatially localized and multi-view consistent.


To address these requirements, we propose \textbf{DesignAgent3D}, an interactive multimodal agentic framework illustrated in Fig.~\ref{fig:Designer_pipeline},   following a \textbf{Plan--Perceive--Act} paradigm 
with a final scene reintegration step.
Concretely, 
first, the \textbf{Planning} step engages the user in an iterative clarification-and-confirmation loop to produce a structured design protocol that specifies the target object and desired appearance. Second, the \textbf{Perception} step uses the object specification from this protocol together with sparse SfM geometry to localize the same physical object across viewpoints and produce view-consistent masks. Meanwhile, it takes the texture prompt from the protocol to generate a unified texture reference, which serves as a global visual guide to keep the object's new appearance consistent across the entire scene. Third, the \textbf{Action} step uses the clarified appearance intent and the resulting multi-view masks to generate edited views with consistent texture and geometry. Finally, these edited views are integrated back into the underlying 3D representation.
Next, we detail the designs of each step.

\begin{figure*}[t]
    \centering
    \includegraphics[width=\textwidth]{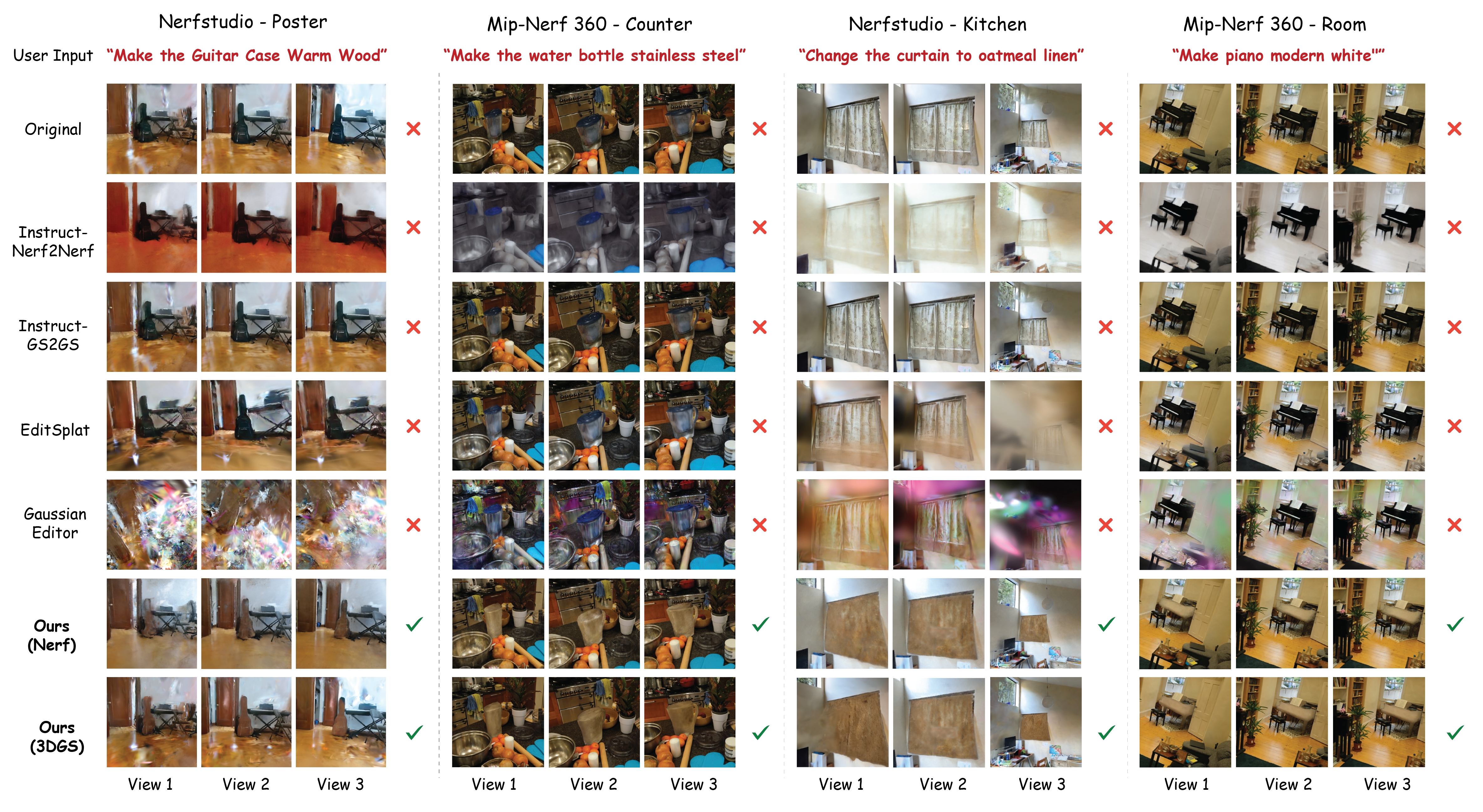} 
    \caption{\textbf{Comparison of 3D Scene Editing Methods.} 
    While existing methods often suffer from global color shifts (e.g., Instruct-NeRF2NeRF~\cite{haque2023instruct}, GaussianEditor~\cite{chen2024gaussianeditor}) or fail to edit any object (e.g., Instruct-GS2GS~\cite{vachha2024instruct}, EditSplat~\cite{lee2025editsplat}), our agentic framework precisely grounds abstract prompts into specific material updates. Moreover, our method maintains sharp object boundaries and superior cross-view consistency, whereas baselines introduce significant "floaters" or blurred textures.} 
    \label{fig:experiment1}
\end{figure*}

\subsection{Planning}

To resolve the semantic ambiguity of natural language design requests, the \textbf{Planning} step converts an initial user request into an executable design protocol through an explicit clarification-and-confirmation loop. Given a user request \(x\), DesignAgent3D does not immediately generate editing prompts. Instead, it maintains a planning state \(h_t\) that records the current specification of the edit, including the target object, desired appearance, material attributes, style preference, and negative constraints. At each interaction turn, the agent    inspects \(h_{t-1}\), identifies missing or ambiguous fields, and asks the user a targeted clarification question   \(q_t\). After receiving the user response \(u_t\), the planning state is updated as:
\begin{equation}
q_t = \pi_{\mathrm{ask}}(x, h_{t-1}), \quad
h_t = \mathrm{Update}(h_{t-1}, q_t, u_t).
\end{equation}
This clarification loop continues until the agent obtains a sufficiently complete specification of \textit{which object} should be edited, \textit{what appearance} should be applied, and \textit{what constraints} should be respected.
The completed planning state is then compiled into a structured design 
\begin{equation}
z = \Phi(x, h_T),
\end{equation}
which serves as the explicit interface between planning and downstream execution.
We implement \(z\) as a compact JSON-like schema containing both object-level and appearance-level instructions. The object field \texttt{<target\_object>} specifies the target instance for downstream segmentation. The appearance fields, including \texttt{<texture\_prompt>}, \texttt{<visual\_prompt>}, \texttt{<inpaint\_prompt>}, and \texttt{<negative\_prompt>}, define the desired material, visual style, repainting instruction, and constraints to avoid undesired artifacts. The object specification is consumed by the \emph{Geometric Anchored Segmentation} module, while the appearance prompts are used by the \emph{Visual Synthesis} module for texture reference generation and view-consistent repainting.

Finally, the generated protocol is shown to the user for confirmation. If the user modifies or rejects the protocol, the agent updates \(h_T\) and regenerates \(z\). Only after confirmation is the protocol passed to the \emph{Perception} and \emph{Action} stages. In this way, \emph{Planning} is not a one-shot prompt rewriting step, but an interactive procedure that constructs, verifies, and freezes the editing intent before any 3D modification is performed.

\vspace{-0.2cm}
\subsection{Perception}

After planning, the \textbf{Perception} step grounds structured design protocol \(z\) into concrete guidance for downstream editing. 
It produces two outputs for the Action stage: view-consistent target-object masks \(\mathcal{M}=\{M_i\}_{i=1}^{N}\) and a user-validated texture reference \(I_{\text{ref}}\).

To obtain these outputs, we design two modules. 
The \textbf{Geometric Anchored Segmentation} module takes the object specification \texttt{<target\_object>} from \(z\) and localizes the same physical target object across different views. 
The \textbf{Texture Reference Generation} module takes the texture-related prompts from \(z\) and produces a user-approved material exemplar for consistent repainting. 
We detail these two modules in Sections~\ref{sec:geometric_segmentation} and~\ref{sec:texture_reference}.

\begin{figure*}[t]
    \centering
    \includegraphics[width=\textwidth]{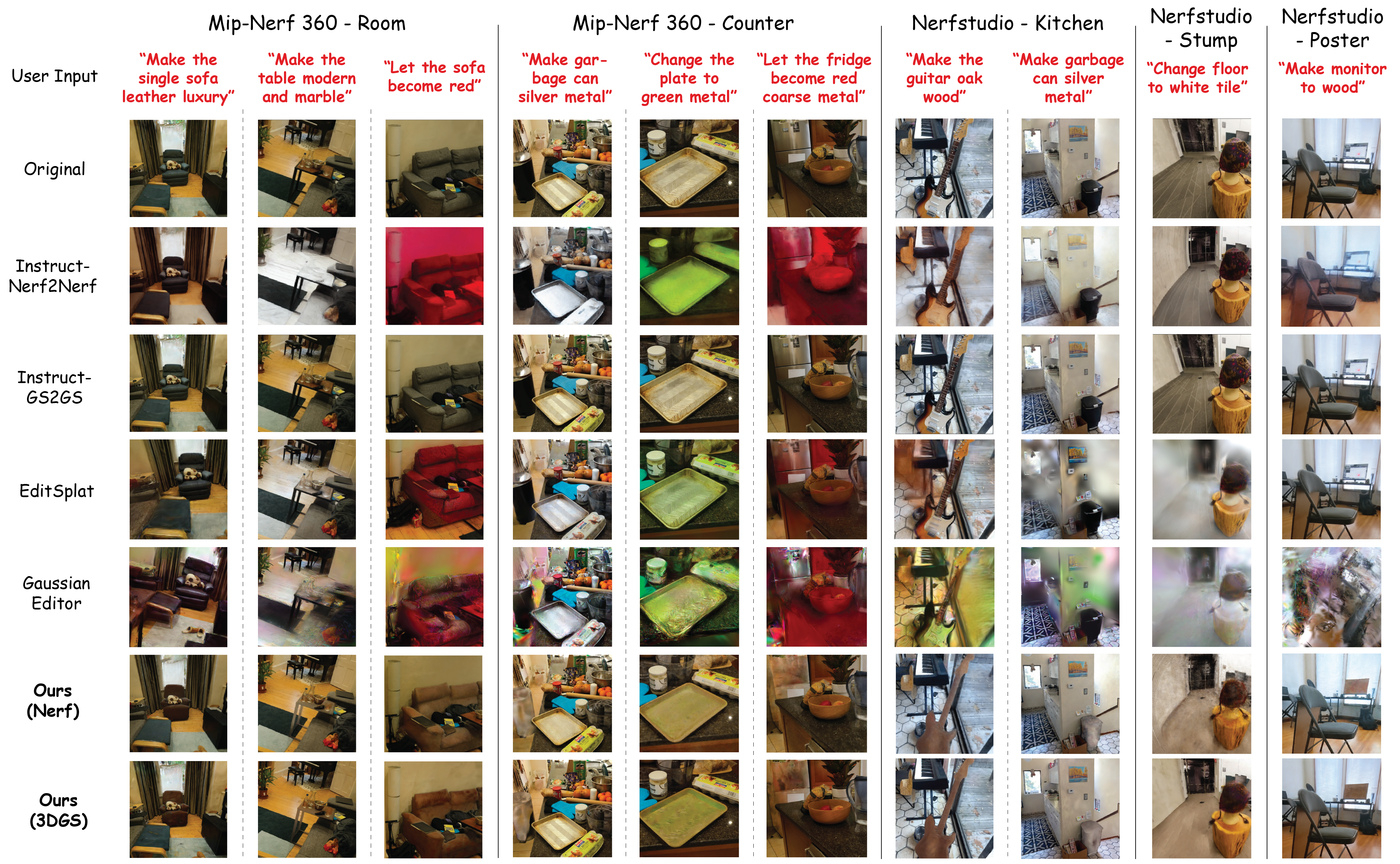} 
    \caption{\textbf{Comparison of 3D Scene Editing Methods (Continue).} Our agentic framework precisely grounds abstract prompts into specific material updates.} 
    \label{fig:experiment2}
\end{figure*}

\subsubsection{Geometric Anchored Segmentation}
\label{sec:geometric_segmentation} 
This module generates view-consistent masks \(\mathcal{M}=\{M_i\}_{i=1}^{N}\) for the target object specified by \texttt{<target\_object>} in the design protocol \(z\). 
As shown in the bottom-left panel of Figure~\ref{fig:Designer_pipeline}, the module consists of two stages: reference segmentation and dataset segmentation.

\noindent\textbf{Stage I: Reference Segmentation.}
We first perform prompt-based object segmentation on candidate views using the clarified \texttt{<target\_object>} as the text cue. 
The segmentation model predicts candidate masks by aligning the visual content with the object description. 
Since a scene may contain multiple similar instances, the predicted masks may still include incorrect or incomplete regions. 
Therefore, the user selects one or a few views where the target object is clearly visible and confirms the correct masks. 
These user-confirmed masks serve as anchor observations for the following 3D grounding step.

\noindent\textbf{Stage II: Dataset Segmentation.}
We then lift the confirmed reference masks into 3D by associating masked pixels with sparse SfM points \(\mathcal{P} \subset \mathbb{R}^3\). 
Specifically, we identify a subset of object proxy points \(\mathcal{P}_{\mathcal{O}}\) that are consistently supported by the confirmed masks across the anchor views:
\begin{equation}
\mathcal{P}_{\mathcal{O}} =
\left\{
\mathbf{X}_k \in \mathcal{P}
\;\middle|\;
\sum_{v \in \{I_a\} \cup \mathcal{V}_{ref}}
\mathbb{1}\!\left[\pi(\mathbf{X}_k,\mathbf{\xi}_v)\in M_v\right]
\ge V_{\min}
\right\},
\end{equation}
where \(\pi(\cdot,\mathbf{\xi}_v)\) denotes the projection function under camera parameters \(\mathbf{\xi}_v\), \(M_v\) is the confirmed mask in view \(v\), and \(V_{\min}\) is the minimum multi-view support threshold. 
This step keeps only the SfM points that are geometrically consistent with the confirmed object masks, yielding a stable 3D proxy representation of the target object.
Once \(\mathcal{P}_{\mathcal{O}}\) is obtained, we project these proxy points into each target view \(j\) to define a constrained search region:
\begin{equation}
B_j = \text{Box}\!\left(
\text{Filter}\!\left(
\left\{
\pi(\mathbf{X}_k,\mathbf{\xi}_j)\mid \mathbf{X}_k \in \mathcal{P}_{\mathcal{O}}
\right\}
\right)
\right),
\end{equation}
where \(\text{Filter}(\cdot)\) removes projection outliers and \(\text{Box}(\cdot)\) constructs the geometric bounding box. 
Within this constrained region, we re-run segmentation to obtain the view-specific object mask while suppressing unrelated background regions and visually similar distractors.

To further reduce mask leakage, we decompose the predicted mask into connected components \(\mathcal{C}=\{c_1,c_2,\dots\}\) and retain the component that receives the strongest support from the projected proxy points:
\begin{equation}
M_j^* = \arg\max_{c\in\mathcal{C}}
\sum_{\mathbf{X}_k \in \mathcal{P}_{\mathcal{O}}}
\mathbb{1}\!\left[\pi(\mathbf{X}_k,\mathbf{\xi}_j)\in c\right].
\end{equation}
The resulting mask \(M_j^*\) is used as the final target-object mask for view \(j\). 
By repeating this process over all views, the module produces \(\mathcal{M}=\{M_i\}_{i=1}^{N}\), which provides spatial guidance for downstream geometry-constrained repainting.

\subsubsection{Texture Reference Generation}
\label{sec:texture_reference}
This module produces a canonical texture reference image \(I_{\text{ref}}\) from the texture-related fields in the design protocol \(z\). 
Specifically, we use \texttt{<texture\_prompt>} as the main text condition to synthesize a 2D material exemplar that specifies the desired color, texture pattern, and surface style. 
This reference image is later used as the shared appearance prior for repainting the target object across different views.

Given the texture prompt \(\mathcal{P}_{\text{tex}}\), we generate the texture reference image as:
\begin{equation}
I_{\text{ref}} = \mathcal{D}\Big(
\text{Denoise}\big(
\epsilon, 
\tau_{\text{L}}(\mathcal{P}_{\text{tex}}), 
\tau_{\text{G}}(\mathcal{P}_{\text{tex}}), 
c_{\text{size}}
\big)
\Big),
\end{equation}
where \(\mathcal{D}\) denotes the pre-trained VAE decoder, and \(\text{Denoise}(\cdot)\) represents the iterative reverse diffusion process. 
The initial noise is sampled as \(\epsilon \sim \mathcal{N}(0,\mathbf{I})\). 
The functions \(\tau_{\text{L}}(\cdot)\) and \(\tau_{\text{G}}(\cdot)\) encode the texture prompt with the dual text encoders of SDXL, and \(c_{\text{size}}\) denotes the size-conditioning vector that controls the resolution and crop information of the generated reference.

After generation, the user reviews the synthesized texture reference. 
If the result does not match the intended material or contains artifacts, the user can either regenerate the reference with a revised prompt or upload an external material image as the reference. 
This review process is repeated until the user approves \(I_{\text{ref}}\). 
The approved texture reference is then fixed and passed to the next Action stage, where it guides view-consistent repainting of the target object.

\subsection{Action}

After Perception, the \textbf{Action} step applies the planned appearance change to the target object and integrates the edited views back into the 3D scene. 
Taking input of  the view-consistent masks \(\mathcal{M}=\{M_i\}_{i=1}^{N}\), the user-validated texture reference \(I_{\text{ref}}\), and the design protocol \(z\),  this step produces two outputs: a set of repainted views \(\mathcal{I}'=\{I'_i\}_{i=1}^{N}\) and an updated 3D scene representation \(\mathcal{S}'\). 
Concretely, the \textbf{Action} is taken by  two stages: \textbf{Geometry-Constrained Repainting}, which edits each view under mask, texture, and depth guidance, and \textbf{Consistent Scene Integration}, which optimizes the 3D representation using the edited views.

\subsubsection{Geometry-Constrained Repainting}

This stage generates an edited image \(I'_i\) for each input view \(I_i\). 
For view \(i\), the repainting model takes four inputs: the original image \(I_i\), the target-object mask \(M_i\), the depth map \(D_i\), and the texture reference \(I_{\text{ref}}\). 
The mask \(M_i\) restricts the editable region, the texture reference \(I_{\text{ref}}\) provides the shared appearance prior, and the depth map \(D_i\) provides geometric guidance for adapting the texture to the local surface structure.

Formally, the repainted image for view \(i\) is modeled as:
\begin{equation}
I'_i \sim \Phi\!\left(
z \;\middle|\;
\mathcal{E}_{\text{img}}(I_i, M_i),
\mathcal{E}_{\text{style}}(I_{\text{ref}}),
\mathcal{E}_{\text{geom}}(D_i)
\right),
\end{equation}
where \(\mathcal{E}_{\text{img}}(I_i, M_i)\) encodes the masked image context, \(\mathcal{E}_{\text{style}}(I_{\text{ref}})\) extracts appearance features from the canonical texture reference, and \(\mathcal{E}_{\text{geom}}(D_i)\) encodes depth-based geometric guidance. 
In implementation, the appearance features are injected through cross-attention, while the depth map is used as ControlNet-style spatial conditioning. 
This conditioning allows the edited texture to follow the local geometry of each view while keeping the same material appearance across views.

\subsubsection{Consistent Scene Integration}

After repainting all views, we obtain the edited image set \(\mathcal{I}'=\{I'_i\}_{i=1}^{N}\). 
We then integrate these edited views into the underlying 3D representation through a replace-and-retrain strategy. 
Specifically, the original training images of the affected views are replaced by their repainted counterparts, and the 3D scene representation is re-optimized to fit the updated observations:
\begin{equation}
\mathcal{S}' = \arg\min_{\mathcal{S}}
\sum_{i=1}^{N}
\mathcal{L}_{\text{rec}}\!\left(
R(\mathcal{S}, \xi_i), I'_i
\right),
\end{equation}
where \(R(\mathcal{S}, \xi_i)\) denotes rendering the 3D scene \(\mathcal{S}\) from camera pose \(\xi_i\), and \(\mathcal{L}_{\text{rec}}\) is the reconstruction loss between the rendered view and the edited image. 
This step embeds the appearance changes into the 3D representation, producing the final edited scene \(\mathcal{S}'\) for consistent novel-view rendering.

\section{Experiments}

\label{sec:experiments}


\begin{figure*}[t]
    \centering
    \includegraphics[width=\textwidth]{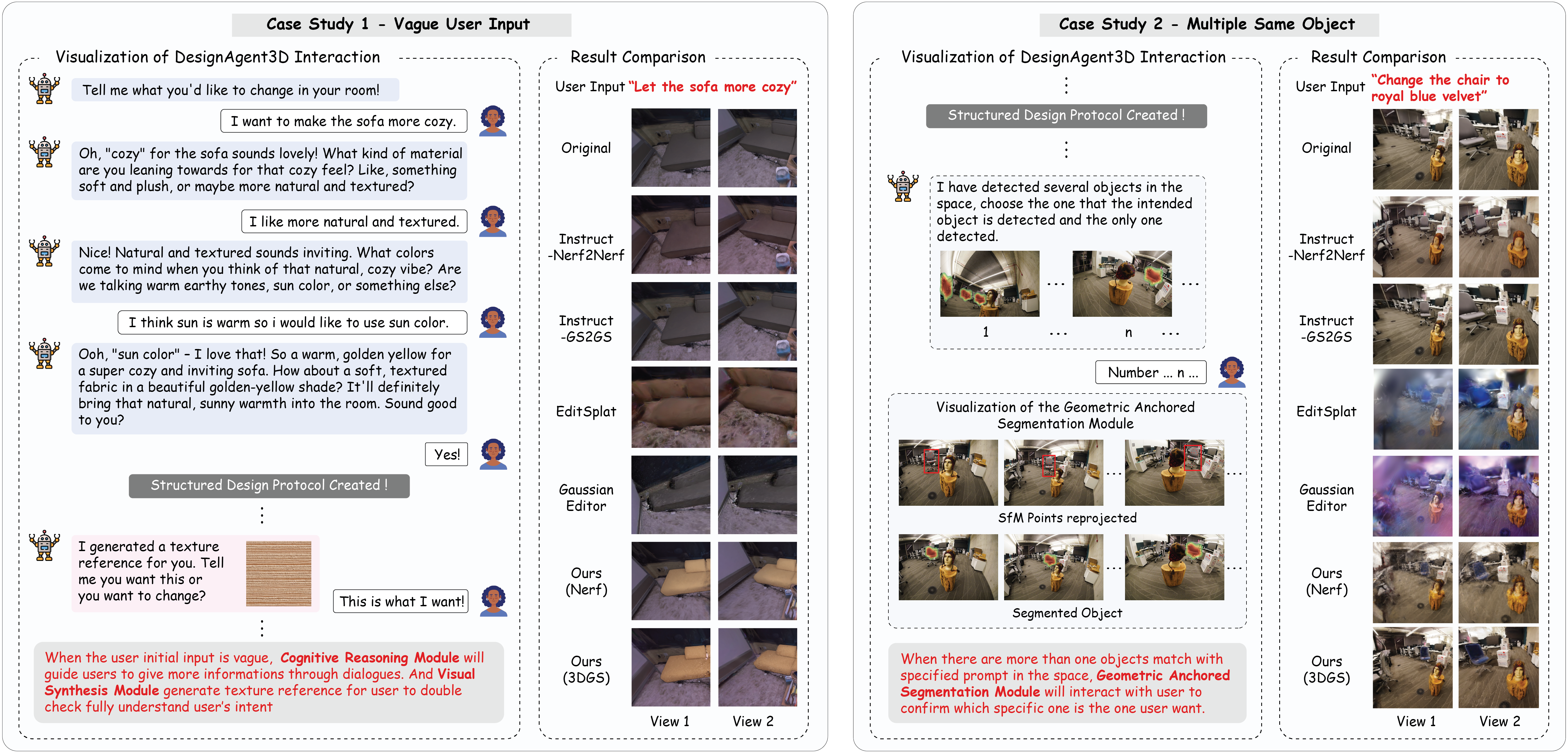} 
    \caption{Case Studies.  
    While baselines "guess" vague intents or fail to distinguish identical objects, our framework uses conversational dialogue and a geometrically anchored segmentation module to clarify goals and confirm target instances via visual candidates.}
    \label{fig:case_study}
\end{figure*}

\subsection{Experimental Setup}

\subsubsection{Datasets and Scenes.}
We evaluate DesignAgent3D on six representative scene-level datasets across three widely used 3D reconstruction benchmarks: \textbf{Nerfstudio}~\cite{mildenhall2021nerf} (\textit{poster}, \textit{kitchen}, \textit{stump}), \textbf{Mip-NeRF 360}~\cite{barron2022mip} (\textit{counter}, \textit{room}), and \textbf{Replica}~\cite{straub2019replica} (\textit{office\_0}). 
Following prior 3D scene editing works~\cite{haque2023instruct, chen2024gaussianeditor, lee2025editsplat}, each scene is treated as an independent evaluation unit, where a separate 3D representation is reconstructed from posed multi-view images before editing. 
In our evaluation, each scene-level dataset contains about 600 posed images on average.

For each scene, we first train an initial 3D reconstruction model, render scene views from the dataset camera poses, perform target-object segmentation and view selection, edit the selected views using the texture reference and inpainter, and finally re-optimize the 3D representation with the edited views. 
This pipeline is repeated independently for each scene-level dataset, so our evaluation measures complete scene-level 3D editing rather than isolated 2D image editing.


\vspace{-0.1cm}
\subsubsection{Baselines.} We evaluate DesignAgent3D against four state-of-the-art 3D editing frameworks: 
(1) \textbf{Instruct-NeRF2NeRF}~\cite{haque2023instruct} establishes an iterative dataset-update scheme, combining 2D image diffusion (InstructPix2Pix) with 3D NeRF optimization for text-driven scene modification. 
(2) \textbf{Instruct-GS2GS}~\cite{vachha2024instruct} adapts this iterative update pipeline to 3D Gaussian Splatting (3DGS) to enable efficient text-driven global edits. 
(3) \textbf{EditSplat}~\cite{lee2025editsplat} incorporates Multi-view Fusion Guidance (MFG) to resolve multi-view inconsistencies and utilizes Attention-Guided Trimming (AGT) on cross-attention weights for precise local editing. 
(4) \textbf{GaussianEditor}~\cite{chen2024gaussianeditor} integrates Gaussian semantic tracing with Hierarchical Gaussian Splatting (HGS) under stochastic 2D diffusion guidance, enabling rapid and stable object removal and integration.

\subsubsection{Evaluation Setup.}
We evaluate DesignAgent3D from both qualitative and quantitative perspectives. 
For qualitative evaluation, we render edited scenes from diverse camera trajectories and compare the visual results with baseline methods. 
For fair comparison, all baselines are provided with the clarified prompts generated by our agent, so that the comparison focuses on each method's editing and reconstruction capability rather than intent clarification.

For quantitative evaluation, following prior 3D editing works~\cite{chen2024gaussianeditor, lee2025editsplat, haque2023instruct}, we use CLIP~\cite{radford2021learning}-based metrics to measure semantic alignment: CLIP Directional Similarity (CLIPdir) measures whether the visual change from the source to the edited image matches the textual change from the source prompt to the target prompt. 
CLIP Text-Image Similarity (CLIPsim) measures the cosine similarity between the edited rendered views and target text prompt.


\vspace{-0.3cm}
\subsubsection{Implementation Details.}
\subsubsection{Implementation Details.}
DesignAgent3D is implemented in PyTorch and primarily built on 3D Gaussian Splatting (3DGS)~\cite{kerbl20233d} for efficient rendering, while also supporting NeRF~\cite{mildenhall2021nerf} to demonstrate backbone adaptability. 
Following EditSplat~\cite{lee2025editsplat}, we execute 1--4 editing tasks per scene, totaling 16 distinct edits. 
Multi-view consistency is validated using rendered videos from unseen viewpoints(see \href{https://anonymous.4open.science/r/DesignAgent3D-F36F}{GitHub repository}).

The Planning module uses Gemini~\cite{team2023gemini} and GPT-4~\cite{achiam2023gpt} APIs to convert user intent into structured design protocols. 
The Perception module uses Grounding DINO~\cite{liu2024grounding} and SAM~\cite{kirillov2023segment} for target grounding and segmentation. 
The Action module employs Stable Diffusion XL~\cite{podell2023sdxl} for localized repainting and performs replace-and-retrain optimization to integrate edited views into the 3D representation.

For each editing session, we run the scene integration optimization for 30,000 iterations, with learning rates of 0.001 for Gaussian positions and 0.01 for color spherical harmonic coefficients. 
All experiments are conducted on a single NVIDIA RTX 3090 GPU with 24GB VRAM.


\begin{table*}[t]
    \centering
    \caption{\textbf{Quantitative Comparison.} CLIP$_{\text{dir}}$: CLIP directional similarity; CLIP$_{\text{sim}}$: CLIP Text-Image Similarity}
    \vspace{-0.4cm}
    \label{tab:quantitative}

    \resizebox{2\columnwidth}{!}{%
        \begin{tabular}{l|cc cc cc cc cc cc cc}
            \toprule

            Scene Edit
            & \multicolumn{2}{c}{Nerfstudio - Poster}
            & \multicolumn{2}{c}{Nerfstudio - Kitchen}
            & \multicolumn{2}{c}{Nerfstudio - Stump}
            & \multicolumn{2}{c}{Mip-Nerf 360 - Room}
            & \multicolumn{2}{c}{Mip-Nerf 360 - Counter}
            & \multicolumn{2}{c}{Replica - Office0}
            & \multicolumn{2}{c}{\textbf{MEAN}} \\

            \cmidrule(lr){2-3}
            \cmidrule(lr){4-5}
            \cmidrule(lr){6-7}
            \cmidrule(lr){8-9}
            \cmidrule(lr){10-11}
            \cmidrule(lr){12-13}
            \cmidrule(lr){14-15}

             & CLIP$_{\text{dir}}$ & CLIP$_{\text{sim}}$
             & CLIP$_{\text{dir}}$ & CLIP$_{\text{sim}}$
             & CLIP$_{\text{dir}}$ & CLIP$_{\text{sim}}$
             & CLIP$_{\text{dir}}$ & CLIP$_{\text{sim}}$
             & CLIP$_{\text{dir}}$ & CLIP$_{\text{sim}}$
             & CLIP$_{\text{dir}}$ & CLIP$_{\text{sim}}$
             & CLIP$_{\text{dir}}$ & CLIP$_{\text{sim}}$ \\

             \midrule

            Instruct-NeRF2NeRF~\cite{haque2023instruct}
             & -0.023 & 0.229
             & 0.020 & 0.266
             & -0.060 & \textbf{0.272}
             & -0.064 & 0.318
             & -0.066 & 0.236
             & 0.061 & 0.246
             & -0.022 & 0.261 \\

            Instruct-GS2GS~\cite{vachha2024instruct}
             & 0.000 & 0.290
             & 0.000 & 0.294
             & 0.000 & 0.263
             & 0.000 & 0.312
             & 0.000 & 0.289
             & 0.036 & 0.238
             & 0.006 & 0.281 \\

            EditSplat~\cite{lee2025editsplat}
             & -0.076 & 0.242
             & -0.003 & 0.248
             & 0.008 & 0.233
             & 0.005 & 0.313
             & 0.004 & 0.315
             & 0.055 & 0.235
             & -0.001 & 0.264 \\

            GaussianEditor~\cite{chen2024gaussianeditor}
             & 0.000 & 0.147
             & -0.007 & 0.243
             & 0.038 & 0.252
             & 0.005 & 0.301
             & -0.003 & 0.276
             & 0.039 & 0.180
             & 0.012 & 0.233 \\

            \midrule

            \textbf{Ours (Nerf)}
             & 0.115 & 0.239
             & \textbf{0.076} & 0.256
             & 0.017 & 0.216
             & 0.027 & 0.314
             & 0.067 & 0.296
             & 0.126 & \textbf{0.306}
             & 0.071 & 0.271 \\

            \textbf{Ours (3DGS)}
             & \textbf{0.177} & \textbf{0.297}
             & 0.062 & \textbf{0.301}
             & \textbf{0.049} & 0.233
             & \textbf{0.069} & \textbf{0.337}
             & \textbf{0.072} & \textbf{0.320}
             & \textbf{0.131} & 0.258
             & \textbf{0.093} & \textbf{0.291} \\

            \bottomrule
        \end{tabular}%
    }
\end{table*}
\vspace{-0.3cm}

\begin{figure}
    \centering
    \includegraphics[width=0.95\columnwidth]{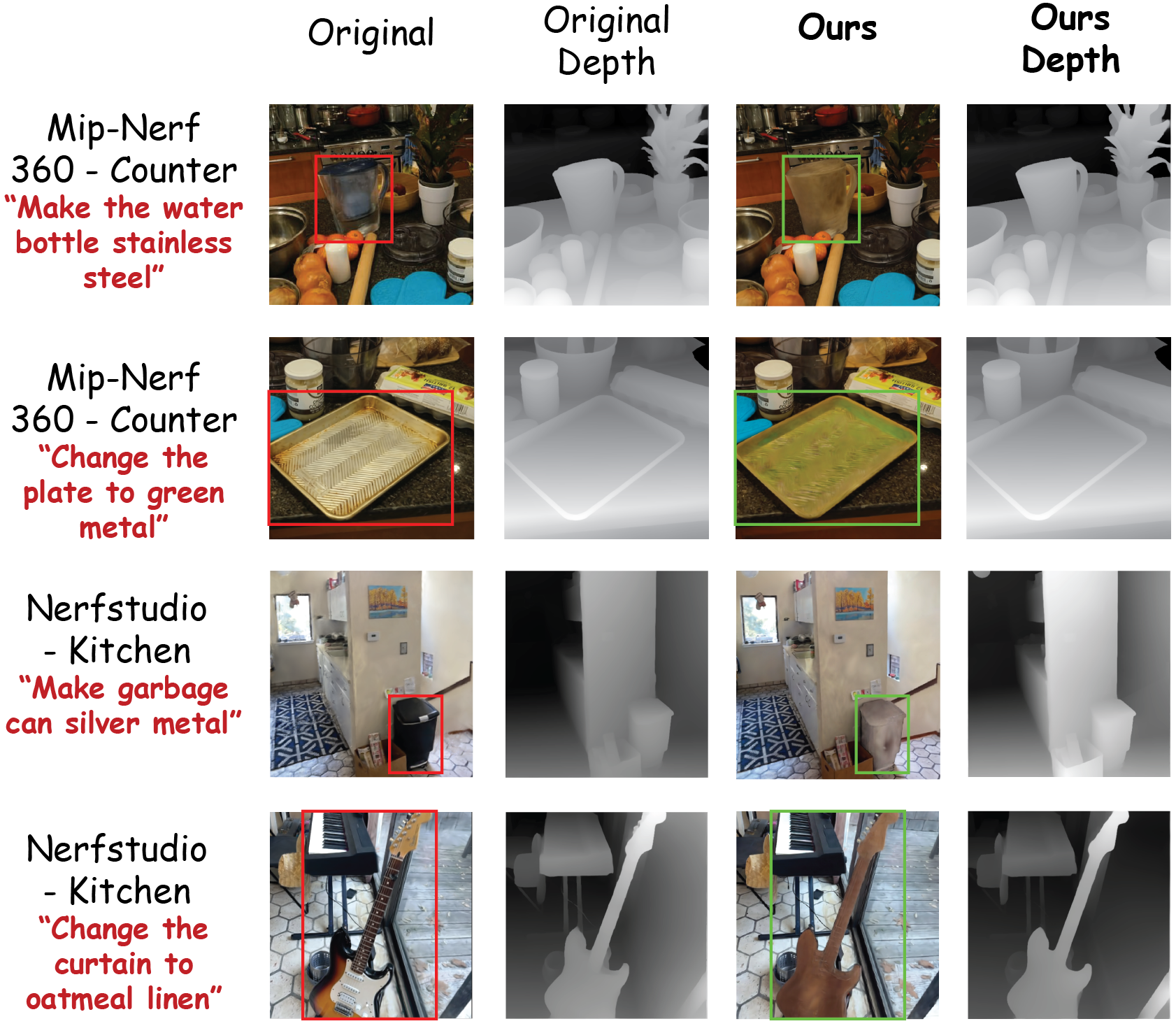} 
    \caption{\textbf{Spatial and geometric integrity evaluation.} We visualize the bounding boxes of target objects before and after editing. Our method maintains a near-perfect IoU, confirming that the editing process does not alter object's original position or spatial footprint within the 3D scene. 
    The high fidelity of the depth maps post-edit proves that our framework effectively preserves the underlying 3D geometry while performing complex texture and material modifications.}
    \label{fig:iou_and_depth}
\end{figure}

\subsection{Qualitative Results}

Figures~\ref{fig:experiment1} and~\ref{fig:experiment2} provide qualitative comparisons from two complementary perspectives. 
Figure~\ref{fig:experiment1} shows representative editing results across three views for each task, highlighting multi-view consistency and object-level localization. 
Figure~\ref{fig:experiment2} presents a broader set of editing tasks with one representative view per task, covering diverse objects, materials, and scenes.

Overall, existing methods exhibit three major failure modes. 
First, Instruct-GS2GS~\cite{vachha2024instruct} often produces nearly unchanged outputs, indicating a ``null-edit'' failure. 
Second, Instruct-NeRF2NeRF~\cite{haque2023instruct} and GaussianEditor~\cite{chen2024gaussianeditor} frequently introduce global color shifts or chromatic artifacts, modifying the background or unrelated regions instead of the target object. 
Third, EditSplat~\cite{lee2025editsplat} sometimes triggers local changes, but the results often contain blurred textures, unstable boundaries, or inconsistent appearances across views.

In contrast, DesignAgent3D produces localized and view-consistent edits across different scenes and materials. 
The improvements mainly come from two design choices: Geometric Anchored Segmentation confines the edit to the intended object, while Texture Reference Generation provides a shared material prior for all views. 
As a result, our method better preserves background regions, maintains sharper object boundaries, and produces more consistent target-object appearances across viewpoints.

We further conduct a case study in Figure~\ref{fig:case_study} to evaluate ambiguous user instructions and scenes with multiple similar objects. 
When the user request is underspecified, DesignAgent3D clarifies the editing intent through dialogue before execution. 
When multiple candidate objects exist, the system uses geometrically anchored segmentation and user confirmation to identify the intended target instance. 
In contrast, baseline methods rely on the raw prompt and often guess the wrong intent or edit the wrong object.

\begin{figure}
    \centering    \includegraphics[width=1\columnwidth]{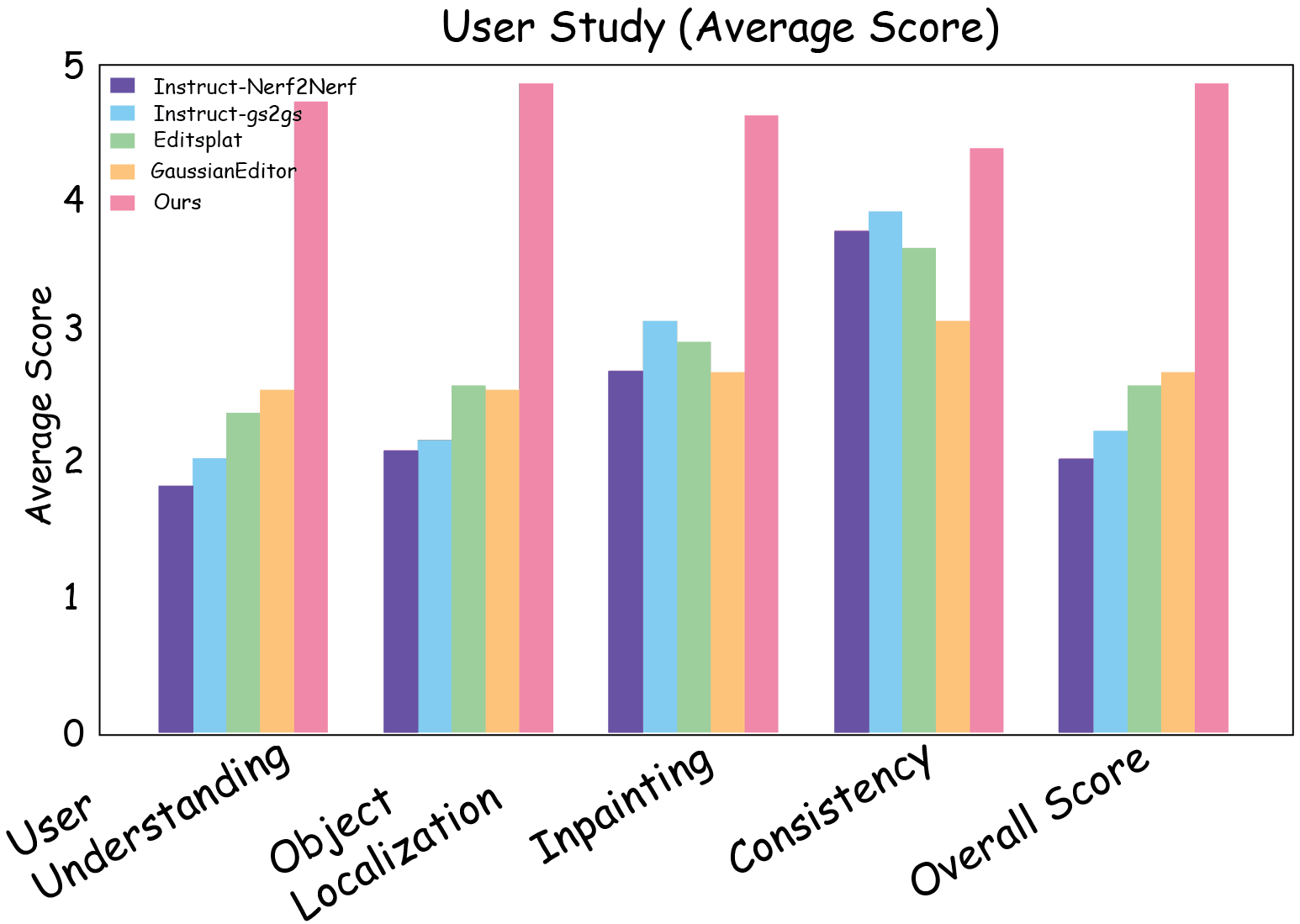} 
    \caption{\textbf{User Study.} The chart compares our framework (Ours) against three baseline models across five key metrics.
    Each bar represents the average score (0–5). Our method consistently outperforms all baselines, demonstrating particular strength in specialized reasoning tasks while maintaining high competitiveness in rendering indicators.}
    \label{fig:user}
\end{figure}
\vspace{-0.2cm}
\subsection{Quantitative Results}

Table~\ref{tab:quantitative} reports the quantitative comparison using CLIP Directional Similarity (\(CLIP_{\text{dir}}\)) and CLIP Text-Image Similarity (\(CLIP_{\text{sim}}\)). 
DesignAgent3D achieves the best mean performance on both metrics. 
For \(CLIP_{\text{dir}}\), our 3DGS-based model obtains a mean score of 0.093, substantially outperforming the best baseline, GaussianEditor (0.012), and improving over EditSplat (-0.001) and Instruct-NeRF2NeRF (-0.022), which even show negative directional alignment on average. 
This indicates that our method produces visual changes that better follow the intended text edit.

For \(CLIP_{\text{sim}}\), our 3DGS-based model also achieves the highest mean score of 0.291, compared with the strongest baseline Instruct-GS2GS (0.281). 
The gain is smaller than that of \(CLIP_{\text{dir}}\), because \(CLIP_{\text{sim}}\) measures global image-text similarity and can be diluted by unchanged background regions in localized 3D editing. 
Nevertheless, our method achieves strong results in challenging scenes such as Mip-NeRF 360 Room and Counter, where accurate target localization and material consistency are required. 
For example, on Room and Counter, our 3DGS-based model achieves the best \(CLIP_{\text{sim}}\) scores of 0.337 and 0.320, respectively, while also obtaining the best \(CLIP_{\text{dir}}\) scores of 0.069 and 0.072. 
This suggests that DesignAgent3D improves target-object editing while preserving the overall scene content.
Moreover, the 3DGS variant outperforms the NeRF variant in mean \(CLIP_{\text{dir}}\) by 31\%, showing that explicit 3D representations provide a more suitable substrate for localized, geometry-aware editing.

\subsection{Spatial and Geometric Integrity}
To evaluate the precision of our editing framework, we investigate whether the modifications strictly adhere to the original scene's spatial layout and geometric structures. A common failure mode in generative 3D editing is "geometric drifting," where the optimization process inadvertently deforms the object's shape or shifts its position while attempting to update its appearance. As illustrated in Figure~\ref{fig:iou_and_depth}, we compare the 2D bounding boxes and Intersection over Union (IoU) between the original and edited objects. Our method achieves an IoU $\approx$ 1 across diverse scenes, demonstrating that the edited target remains perfectly anchored within its original spatial boundaries. Furthermore, we analyze the structural consistency through depth map comparisons. As shown in Figure~\ref{fig:iou_and_depth}, the depth maps of the edited scenes remain nearly identical to the original ones. This indicates that our geometrically anchored segmentation and multi-agent coordination successfully decouple appearance editing from geometric reconstruction. By preserving the underlying geometry, our method ensures that the edited objects integrate seamlessly into the surrounding environment without introducing physical distortions or spatial inconsistencies.
\vspace{-0.1cm}
\subsection{User Studies}
To account for human perception, we conducted a blind user study with 15 professional practitioners from architecture, design, and engineering. 
Users were asked to rank our method against baselines based on four criteria: (1) adherence to user intent, (2) object localization, (3) inpainting quality, and (4) view consistency.
We evaluate five key metrics: User Understanding, Object Localization, Inpainting Quality, Consistency, and Overall Score. 

As illustrated in Figure~\ref{fig:user}, baseline models show clear performance drops in high-level reasoning metrics, particularly User Understanding and Object Localization. 
This mainly stems from their inability to resolve spatial ambiguities in complex natural language prompts, leading to object-level hallucinations where edits are applied to incorrect regions.
In contrast, our framework achieves stronger User Understanding and Object Localization by combining the agentic clarification loop with geometric anchoring, ensuring that the intended edit is grounded to the correct physical object. 
For Inpainting Quality and View Consistency, our method also achieves competitive results by using texture-guided repainting and multi-view constraints to reduce identity drift and flickering artifacts. 
These findings demonstrate that our framework better bridges abstract instructions and physical target objects, while maintaining competitive rendering quality for object-level 3D scene editing.

\begin{figure}
    \centering
    \includegraphics[width=1\columnwidth]{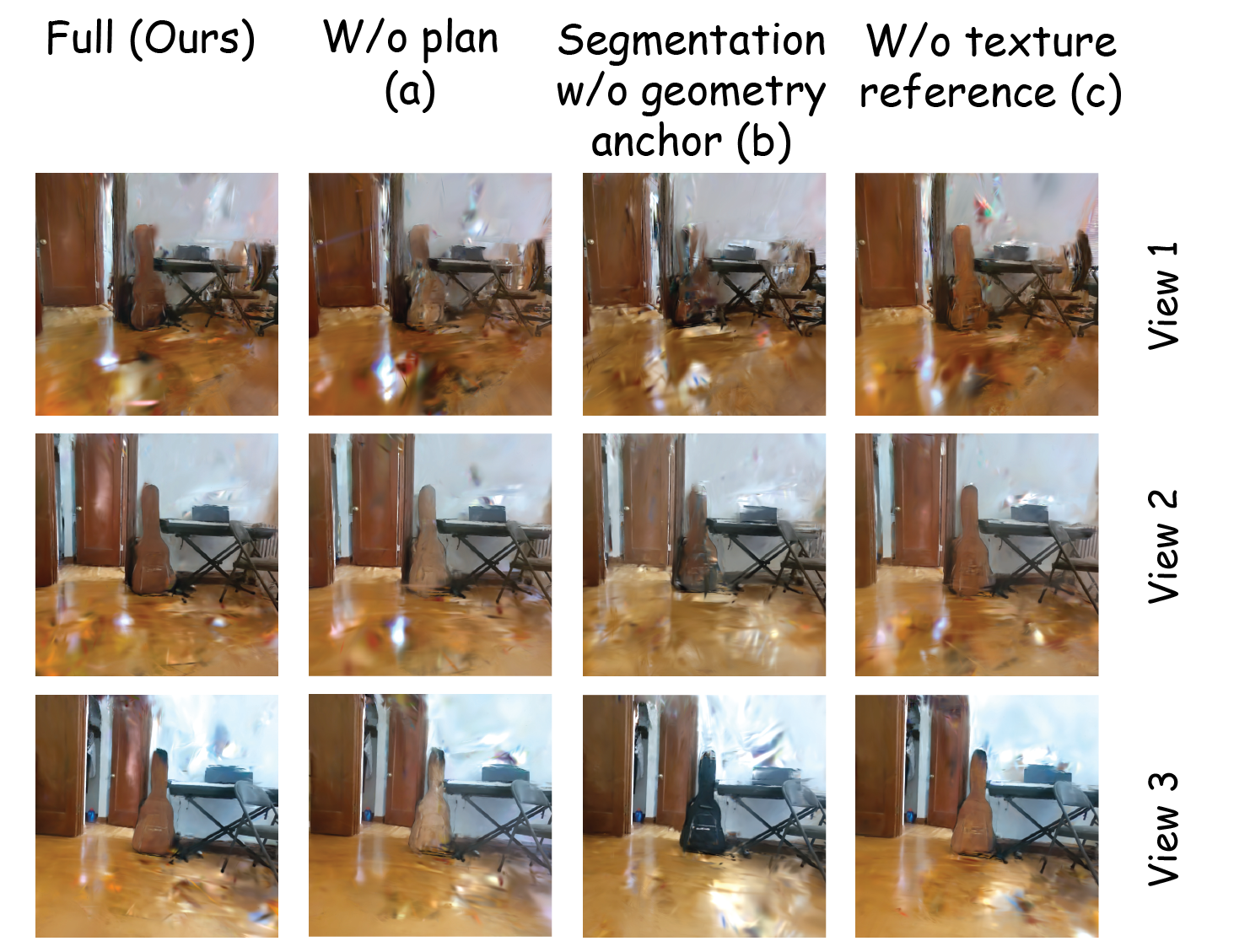} 
    \caption{\textbf{Ablation Studies.} Evaluate the necessity of our core components: (a) Plan: Without interactive loop, the model fails to ground abstract concepts, resulting in irrelevant style shifts. (b) Geometry Anchored Segmentation: Replacing our 3D-aware module with 2D-only methods leads to inaccurate masks (e.g., over-exclusion) and floaters. (c) Texture Reference: Our dual-stream strategy ensures superior cross-view consistency compared to single-prompt baselines, which suffer from flickering and appearance drifting.}
    \label{fig:ablation}
\end{figure}
\vspace{-0.1cm}
\subsection{Ablation Studies}

We conduct ablation studies to verify the necessity of our core modules as summarized in Figure \ref{fig:ablation}.

\noindent
\textbf{The Effectiveness of Cognitive Reasoning.} We compared our interactive agentic guidance loop against a single shot baseline where the LLM directly guesses the intent from the raw user prompt. As shown in the qualitative comparison in Figure \ref{fig:ablation}(a), without the structured discovery loop, the generative models cannot ground abstract design concepts into concrete visual attributes, leading to stylistic ambiguity or irrelevant style shifts.

\noindent
\textbf{The Effectiveness of Geometry Anchored Segmentation.} We replace our Geometry Anchored Segmentation with purely 2D semantic methods. As illustrated in Figure \ref{fig:ablation}(b), the absence of rigid geometric priors leads to unstable segmentation behavior and compromised detection accuracy. Specifically, this leads to two common failure modes: over-exclusion, where the model overlooks valid target objects, and false positives, where unrelated background elements are incorrectly included. This lack of 3D grounding not only results in fragmented masks and temporal artifacts (e.g., flickering and floaters) but also introduces semantic ambiguity, making it difficult to isolate a specific target when multiple instances of the same category coexist.

\noindent
\textbf{The Effectiveness of Texture Reference.} As shown in \ref{fig:ablation}(c), we observe that using a single prompt without a texture reference primarily compromises multi-view consistency. Without the grounding provided by a texture prior, the model struggles to maintain a coherent appearance across different viewpoints, leading to visible flickering and appearance drifting. Our Dual-Stream Prompting strategy successfully decouples these properties, ensuring that material updates remain view-consistent while preserving the original object silhouette with high precision.

\section{Conclusion}

We presented DesignAgent3D, a multimodal agentic framework for interactive object-level indoor 3D scene editing that couples intent clarification with geometry-aware execution. 
DesignAgent3D first translates ambiguous user requests into a structured design protocol, then grounds the target object with SfM-guided Geometric Anchored Segmentation, and finally synthesizes edits through texture-referenced, geometry-constrained repainting. 
The replace-and-retrain integration step further embeds the edited views into the underlying 3D representation, enabling persistent and view-consistent scene updates. 
Experiments, user studies, and ablations across multiple indoor scenes validate the effectiveness of each component and demonstrate consistent advantages over strong 3D editing baselines.
\bibliographystyle{ACM-Reference-Format}
\bibliography{sample-base.bib}
\end{document}